\def\year{2021}\relax
\documentclass[letterpaper]{article} 
\usepackage{aaai21}  
\usepackage{times}  
\usepackage{helvet} 
\usepackage{courier}  
\usepackage[hyphens]{url}  
\usepackage{graphicx} 
\usepackage{natbib}  
\usepackage{caption} 
\usepackage{cite}
\usepackage{amsmath,amssymb,amsfonts}
\usepackage{algorithmic}
\usepackage{graphicx}
\usepackage{textcomp}
\usepackage{xcolor}
\usepackage{longtable}
\usepackage{booktabs}
\usepackage[algo2e,ruled]{algorithm2e}
\usepackage{textcomp}
\usepackage{xcolor}
\usepackage{mathrsfs,enumitem}
\usepackage{mathtools}
\usepackage{subcaption}
\usepackage[export]{adjustbox}
\usepackage[draft]{todonotes}
\usepackage{multirow}
\title{Improving Adversarial Robustness in Weight-quantized Neural Networks}
\author{
    Chang Song\textsuperscript{\rm 1}, Elias Fallon\textsuperscript{\rm 2}, Hai Li\textsuperscript{\rm 1}
    \\
}
\affiliations{
    \textsuperscript{\rm 1}Duke University\\

    2080 Duke University Road,\\
    Durham, NC 27708\\
    \textsuperscript{\rm 2}Cadence Design Systems, Inc.\\
    Allison Park, PA\\
    chang.song@duke.edu, fallon@cadence.com, hai.li@duke.edu

}

\begin{document}

\maketitle

\begin{abstract}
Neural networks are getting deeper and more computation-intensive nowadays. Quantization is a useful technique in deploying neural networks on hardware platforms and saving computation costs with negligible performance loss. 
However, recent research reveals that neural network models, no matter full-precision or quantized, are vulnerable to adversarial attacks. 
In this work, we analyze both adversarial and quantization losses and then introduce criteria to evaluate them. 
We propose a boundary-based retraining method to mitigate adversarial and quantization losses together and adopt a nonlinear mapping method to defend against white-box gradient-based adversarial attacks. 
The evaluations demonstrate that our method can better restore accuracy after quantization than other baseline methods on both black-box and white-box adversarial attacks. 
The results also show that adversarial training suffers quantization loss and does not cooperate well with other training methods.
\end{abstract}

\section{Introduction}
\label{intro}

Neural networks achieve rapid development recently and have been widely applied in various areas~\cite{lecun2015deep}. 
With more layers and more complex structures, neural network models demonstrate near or even beyond human-level accuracy in classification problems~\cite{graham2014fractional}. Neural networks not only become main trend in image recognition, object detection, speech recognition, and natural language processing tasks, but also are widely adopted by security industry~\cite{tao2006human}, including surveillance, authentication, facial recognition, vehicle detection, and crowd behavior analysis.

With the assist of rapid development of hardware platforms, neural network architecture evolve dramatically, handling more complicated datasets and achieving better performance. 
From initially shallow structures like LeNet~\cite{lecun1998gradient} and AlexNet~\cite{krizhevsky2012imagenet}, to deeper and wider designs such as GoogLeNet (a.k.a Inception)~\cite{szegedy2015going}, VGG~\cite{simonyan2014very} and ResNet~\cite{he2016deep}, the parameters of neural network models grow exponentially from 60 thousand to over 100 million, and the total multiply-accumulate operations (MACs) increases from 341 thousand to 15 billion. 
Such explosions in model sizes and MACs require large computation resources and memory storage and result in high power consumption.
This makes the deployment of neural networks very challenging, especially on low-power mobile devices, such as autonomous vehicle systems~\cite{maqueda2018event} and real-time robot navigation systems~\cite{ribeiro2017real}. 
In order to fit neural networks into such resource-constrained hardware platforms, model compression methods such as quantization~\cite{hubara2017quantized}, sparsification~\cite{NIPS2016_6504,liu2015sparse}, and pruning~\cite{han2015learning} have been extensively studied to save memory and computation costs, improve execution speed, while maintaining performance.

In addition to the execution efficiency, the deep learning research community pays a close attention to security and privacy problems in the last few years. 
Recent research shows that neural network models are vulnerable to certain attacks, especially the so-called adversarial attacks~\cite{szegedy2013intriguing}. 
Adversarial attacks can easily fool models and lead to misclassifications by elaborately generating imperceptible perturbations adding to inputs (a.k.a. adversarial examples). 
Moreover, adversarial attacks are transferable, that is, adversarial examples generated by a model can even influence other models with different structures~\cite{papernot2016transferability}. 
Such imperceptible attacks with great transferability unveil the blind spots in neural networks and raise severe reliability and security concerns.
As the research on machine learning and neural networks go deeper, different types of adversarial attacks are proposed.
However, most defenses are targeted and effective only to one specific attack. 
For example, Athalye \emph{et al.} claim that the attacks based-on the obfuscated gradients they proposed could circumvent 7 out of 9 noncertified white-box-secure defenses that were accepted by the \textit{International Conference on Learning Representations (ICLR)} in that year~\cite{athalye2018obfuscated}. 
To better address the universality issue and to fully analyze adversarial attacks, some researchers focus on robustness of models instead of solely relying on accuracy of certain attacks.

In this work, we aim to improve the robustness of weight-quantized neural networks. 
Compared to existing works on defending adversarial attacks and improving neural network robustness, our major contributions include: 
\begin{itemize}[leftmargin=*]
    \item We theoretically analyze both adversarial and quantization losses and come up with using the Lipschitz constant to measure the quantization error amplification effect in each neural network layer.
	\item We observe that adversarially-trained neural networks are vulnerable to quantization loss, which means adversarial training can not be directly used in quantized models.
	\item We deploy a boundary-based error-tolerant retraining method to minimize both losses at the same time and adopt a nonlinear mapping method to defend white-box gradient-based attacks.
	\item We validate the effectiveness of the proposed method through extensive experiments and evaluations. Results also indicate that adversarially-trained models do not cooperate well with other training techniques.
\end{itemize}

The remainder of this paper is organized as follows: 
\textit{Background} summarizes the background about adversarial attacks, model robustness, and the  quantization process. 
\textit{Motivation and Theoretical Explanation} analyzes both adversarial and quantization losses and finds that adversarial models are more vulnerable to quantization loss. 
\textit{Methodology} describes the details of all the procedures to improve adversarial and quantization robustness. 
\textit{Experimental Evaluations} presents and discusses experimental results. 
In the end, we conclude this work in \textit{Conclusions}.
\section{Background}
\label{background}

\subsection{Adversarial Examples and Attacks}

Adversarial attacks generate adversarial examples to fool models. 
An adversarial example $\widetilde{X}$ is generated by injecting adversarial perturbation $\epsilon$ to a clean sample $X$, such as: $\widetilde{X}=X+\epsilon$.
Usually, adversarial perturbations are so tiny that are even imperceptible to human eyes. 
However, the carefully designed adversarial perturbations can fool a neural network to misclassify adversarial examples with high confidence levels~\cite{goodfellow2014explaining}. 
There are multiple methods to generate $\epsilon$, such as FGSM~\cite{goodfellow2014explaining}, CW~\cite{carlini2017towards}, and PGD~\cite{madry2017towards}. $\epsilon$ is also referred to as adversarial strength, larger $\epsilon$ means $\widetilde{X}$ is further away from $X$ in the decision space. 
Adversarial attacks raises security concerns about the applications of neural networks in real-world scenarios. 
For example, attackers could alter a ``stop" sign imperceptibly and make an autonomous vehicle interpret as a ``yield" sign~\cite{papernot2017practical}.

According to the attackers' knowledge, adversarial attacks can be classified as \textit{black-box} attacks and \textit{white-box} attacks. 
Black-box attacks generally mean that the attackers can access only the inputs and outputs of the target model. They don't have information about the target model's structure and parameters. 
Thus the attackers usually design and train their own model to generate adversarial examples. 
In white-box attacks, the attackers have a full access to the target model, and therefore can elaborate specific adversarial examples against the target model.
A given model usually has worse performance against white-box attacks than black-box attacks (with the same attack strengths), as the attacker has more knowledge about the model in white-box scenarios. 
Due to the model transferability~\cite{papernot2016transferability}, black-box attacks are feasible but they usually need larger adversarial strengths to downgrade accuracy compared to white-box attacks. 

Other key adversarial attack definitions include \textit{targeted} vs. \textit{non-targeted} attacks, \textit{iterative} vs. \textit{non-iterative} attacks~\cite{yuan2019adversarial}. 
The goal of targeted attacks is to fool a model to classify examples in one class as a specific class, while non-targeted attacks aim to increase the model's misclassification rate but do not care the classification results. 
Iterative attacks perform multiple steps of attacks to optimize the final adversarial examples and non-iterative attacks only perturb the samples once.

In this paper, we mainly focus on non-targeted attacks. And we will give equal consideration to white-box and black-box, iterative and non-iterative attacks.

\subsection{Robustness}

Robustness indicates how well models can tolerate different types of perturbations and remain proper function. 
Before adversarial attacks were discovered, robustness is discussed mainly in random noise scenarios~\cite{ulivcny2016robustness}. 
Adversarial perturbations $\epsilon$ can be regarded as the worst-case noise which result in large classification errors with minimum efforts. 
In this work, we consider accuracy of different types of test examples as an indicator of robustness: \textit{if a model can achieve an overall high accuracy against multiple types of attacks, the model is robust.}

So far, adversarial training is taken as the most effective and efficient way to mitigate the damage of adversarial attacks and to improve model robustness.
It includes adversarial examples in the training set to enhance models’ resilience to adversarial attacks. 
The main idea of adversarial training is similar to data augmentation: to facilitate training by providing more information on data distribution. 
The effectiveness of adversarial training has been demonstrated and explained by~\cite{goodfellow2014explaining}. 
Note that in adversarial training, the model that is used to generate the adversarial examples is not necessarily identical to the model being attacked.

Some other popular methods have been studied to improve the robustness of neural networks. 
Generally, these methods can be cataloged as two types: (1) Gradient masking: Its main idea is to build a model to hide or smooth the gradient between original and adversarial examples~\cite{papernot2017practical}.
It is particularly effective against white-box attacks. When the attacker use a model different from the protected model, the performance can significantly degrade.
(2) Defensive distillation: The target of this method is to create a model whose decision boundaries are smoothed along the directions that the attacker may exploit. Defensive distillation makes it difficult for the attacker to discover adversarial input tweaks that lead to incorrect classes.

\subsection{Quantization}

The recent performance enhancement of neural networks mainly benefits from more parameters and operations. Such costs are not tolerable for limited-power or real-time applications. 
The neural network quantization helps save computation and memory costs and therefore improve power efficiency. 
Moreover, quantization enables the deployment of neural networks on limited-precision hardware. 
Many hardware systems have physical constraints and do not allow full-precision models directly mapping to them. 
For example, GPUs support half precision floating point arithmetic (FP16), while using lower bit widths gives significant speedups and saves energy consumption. ReRAM (a.k.a memristor)~\cite{strukov2008missing} only allows limited precision because of process variations~\cite{6709674,10.1145/2744769.2744900}.

There are two types of quantization methods: \textit{post-training quantization} and \textit{quantization-aware training}. 
Post-training quantization directly quantizes a model without retraining. However, recent research show that 8-bit post-training quantization does not preserve accuracy in some cases, especially for small models~\cite{jacob2018quantization}. 
Quantization-aware training emulates inference-time quantization, and keeps a copy of full-precision weights to accumulate gradient changes. 
During the backpropagation, the quantization functions are discrete-valued and cannot produce derivatives for training. 
A popular bypass is using the straight-through estimator (STE) to approximate the backpropagation process~\cite{bengio2013estimating}.

The quantization levels could be linear or logarithmic~\cite{miyashita2016convolutional}. 
In linear quantization, quantization levels are uniformly distributed. While in logarithmic quantization, quantization levels follow logarithmic functions. 
Both of them are feasible and show similar effect.
In this work, we consider only the linear distribution for simplicity. 
The quantization can be applied to weight parameters as well as activations. 
It has been observed that activations are more sensitive to quantization than weights~\cite{zhou2016dorefa}. 
Some works quantize activations to a higher precision than weights, some only quantize weights and keep activation at full-precision, and only a few studies explore both~\cite{8999057}. 
We consider the weight-quantization only as we agree that quantization do more harm to activations than to weights. 

Recent works show that quantization can defend adversarial attacks to some extent~\cite{galloway2017attacking}, acting as a gradient masking method. We believe both neural network security issues and hardware implements are crucial and it is the reason that we further explore robustness in quantization scenarios based on existing research.
\section{Motivation and Theoretical Explanation}
\label{motivation}

\subsection{Adversarial and Quantization Losses}

For a single fully-connected layer, suppose $W$ is the weight matrix, $W+\Delta W$ is the weight after quantization, $x$ is the original input, and $x+\Delta x$ is the adversarial input. The difference in the output of this layer ($\delta$) can be represented as followed:
\begin{equation}
    \label{eq:1}
    \delta = (W+\Delta W) \cdot (x+\Delta x)-Wx = \underbrace{W\Delta x}_{\text{Adv.}} + \underbrace{\Delta Wx}_{\text{Quant.}} + \Delta W\Delta x.
\end{equation}
The first term on the right hand side of Eq.~(\ref{eq:1}) is the adversarial loss introduced by adversarial examples, the second term is the quantization loss introduced by weight quantization, and the third term is the loss introduced by adversarial examples and quantization together. Since the third term is much smaller than the first two, we omitted it for simplicity.

The adversarial loss can be easily measured by the accuracy drop under adversarial attacks. For quantization loss, the accuracy drop could reflect the change in weights, but not directly. 
The difference in outputs between a full-precision model and its quantized counterpart reflect the effect of quantization, but it also depends on the input. 
Thus, an input-independent criterion is essentially necessary to evaluate the quantization process.

\subsection{The Quantization Error Amplification Effect and the Lipschitz Constant}

The quantization error amplification effect was initially introduced in~\cite{liao2018defense}. 
It indicates the situation that small residual perturbation is amplified to a large magnitude in top layers of a model and finally leads to a wrong prediction. 
This effect mentioned in~\cite{liao2018defense} mainly focuses on the input-level perturbation, especially adversarial perturbations. 
In this work, we focus more on the weight-level error. 
Similar to its original definition, the quantization error amplification effect means that each layer will inject a quantization loss. 
The error introduced by quantization can be amplified layer by layer and eventaully results in misclassification in the output.

To measure the influence of quantization loss, we propose to use the Lipschitz constant as a quantization error amplification indicator. 
Based on~\cite{cisse2017parseval}, the Lipschitz constant of the weight loss ($\Delta W$) introduced by quantization can be defined as:
\begin{equation}
    \label{eq:2}
    \left \| \Delta W \right \|_p = \sup_{z: \left \| z \right \|_p = 1} \left \| \Delta Wz \right \|_p ,
\end{equation}
where, $z$ is an arbitary normalized input. 
Eq.~(\ref{eq:2}) shows that the quantization loss can be upper-bounded by the Lipschitz constant of $\Delta W$. 
A Lipschitz constant greater than 1 denotes that the weight undergoes a large quantization loss, otherwise a Lipschitz constant less than 1 means the weight has a small loss after quantization.

$\left \| \Delta W \right \|_2$ when $p=2$ is called the spectral norm of $\Delta W$, which is the maximum singular value of $\Delta W$. 
$\left \| \Delta W \right \|_\infty$ when $p=+\infty$ is the maximum 1-norm of the rows of $\Delta W$. Thus, we have a quantitative way to measure quantization loss using the Lipschitz constant.

All the layers aforementioned in \textit{Motivation and Theoretical Explanation} are fully-connected layers, which perform basic matrix-vector or matrix-matrix multiplications. For more complicated layer structures, such as convolutional layers, \cite{cisse2017parseval} has already given detailed explanations on how to represent convolution operations as basic matrix multiplications and we have verified the correctness of their derivations. 

\subsection{Adversarial Training and Quantization}

\begin{figure*}[ht]
	\centering
	\includegraphics[width=\linewidth]{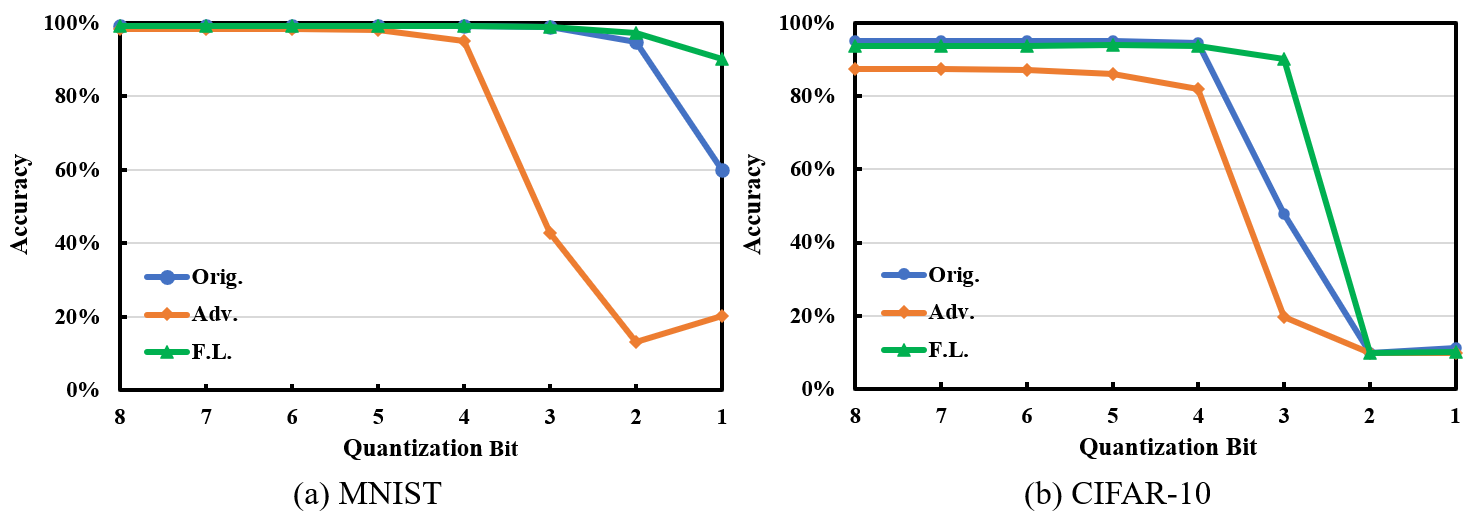}
	\caption{Clean Accuracy vs. Quantization Levels.}
	\label{fig:motivation}
\end{figure*}

As discovered in \cite{lin2019defensive}, quantized models are more vulnerable to adversarial attacks than their original full-precision models. 
We reproduce their experiments and obtained the similar results. 
Furthermore, we find that adversarially-trained models are vulnerable to the quantization process.
Figure~\ref{fig:motivation} presents two example cases (refer \textit{Experimental Evaluations} for the setup details).
Here, Orig. is a vanilla model with standard training methods, Adv. was adversarially-trained by~\cite{madry2017towards}, F.L. was retrained from the original model using feedback learning, as proposed in~\cite{8999302}.

Figure~\ref{fig:motivation} shows that for both examples, as the weight quantization level decreases, the accuracy of adversarially-trained models drops more quickly than that of other models. 
For both MNIST and CIFAR-10 datasets, the F.L. model has the best resistance to the weight quantization. 
The adversarial model is vulnerable to quantization loss, which has an even lower accuracy than the original model, especially when the precision level is low. 
For the MNIST dataset, the accuracy gap between the adversarial model and the other two models enlarges as the quantization level drops. 
In contrast, for the CIFAR-10 dataset, all three models depreciate at 1-bit and 2-bit quantization. 
The different performance of two datasets implies that complicated datasets and models need higher precision than simple datasets and models. 
In this work, we mainly focus on 3-bit quantization to get better a understanding of different training techniques.

Here is a possible explanation of why adversarial training does not perform well in the quantized model. 
Adversarial training introduces additional examples to the training set, which helps reshape decision boundaries to better defend adversarial attacks. However, it fails to compensate for the quantization effect on decision boundaries and even leads to larger quantization loss. 
We will validate this explanation in \textit{Experimental Evaluations} using Lipschitz constant.
\section{Methodology}
\label{method}

Our proposed method integrates two orthogonal components 
to improve overall robustness against adversarial and quantization losses, 
including a boundary-based error-tolerant retraining method to minimize both adversarial loss and quantization loss, and a nonlinear mapping method to defend white-box adversarial attacks. 
In this section, we will elaborate the details of the proposed method.



\subsection{Boundary-based Error-tolerant Retraining}

To minimize both adversarial loss and quantization loss at the same time, we seek for a method that can utilize the similarity between these two types of losses. 
Adversarial loss occurs as adversarial examples move away from original data points in the decision space and cross the model's decision boundaries. 
By adding adversarial examples into the training set, adversarial training can slightly relocate the decision boundaries to reduce adversarial loss. 
Quantization loss is introduced when weights are quantized to discrete values, which also alter the location of decision boundaries. 
Thus, a direct intuition is to find a boundary-based method that can minimize both losses. 

Feedback learning proposed by~\cite{8999302} measures the boundary information of a pre-trained model and then generates examples to increase the margins between samples and decision boundaries in the decision space. 
It is similar to adversarial training w.r.t. data augmentation, but more generalized than adversarial training. 
It can efficiently improve adversarial robustness and also increase tolerance between data and decision boundaries. 
In our work, we leverage feedback learning and take advantage of its margin-improving ability to save quantization loss.

\subsection{Defending White-box Gradient-based Attacks with Nonlinear Mapping}
\begin{figure}[h]
	\centering
	\includegraphics[width=\linewidth]{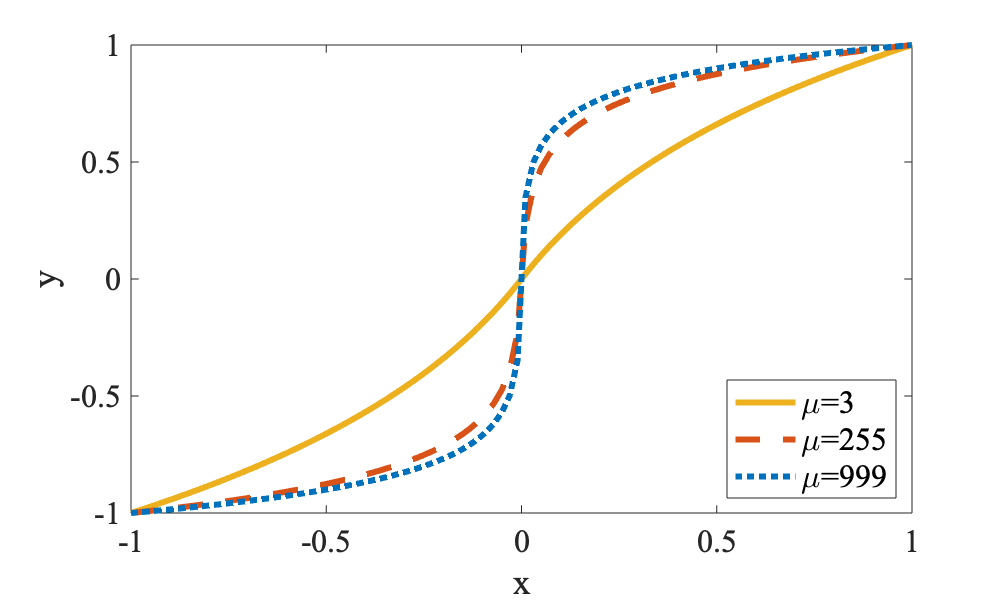}
	\caption{An illustration of $\mu$-law nonlinear mapping.}
	\label{fig:mu-law}
\end{figure}

As previously explained in \textit{Background}, white-box attacks may have more critical influence than black-box attacks with the same attack strengths. 
Here we introduce model nonlinearity as a gradient masking method to improve model's robustness against white-box gradient-based attacks. We adopt a nonlinear mapping method called $\mu$-law algorithm used in wireless communication. 
$\mu$-law is a compressing-expanding (a.k.a. companding) algorithm that compresses the signals before transmitting on a band-limited channel and then expands them when received.
The process can be formulated by 
\begin{subequations}
\begin{align}
    F(x) & =sgn (x)\frac{\ln (1+\mu \left | x  \right |)}{\ln (1+\mu )},~-1 \leq x \leq 1,\text{~and}\label{eq:3}\\
    F^{-1}(y) & =sgn (y)\frac{(1+\mu )^{\left | y \right |}-1}{\mu},~-1 \leq y \leq 1. \label{eq:4}
\end{align}
\end{subequations}
Where, $x$ is the input signal, $y$ is the companded signal, $sgn(\cdot)$ is the sign function. 
Eq.~(\ref{eq:3}) is for compression, and Eq.~(\ref{eq:4}) is for expansion. 
We will only use the compression function to increase nonlinearity. 
Fig.~\ref{fig:mu-law} shows that the curvatures and nonlinearities of the compression mapping curves increase when $\mu$ increases.
In other words, by controlling the value of $\mu$, we can control the nonlinearity of a model after mapping.

In analog systems, this method can increase the signal-to-noise ratio (SNR) achieved during transmission. In the digital domain, it can reduce the quantization error, and hence, increase the signal to quantization noise ratio. After neural network training, mapping the weights directly to a different distribution 
can increase model nonlinearity. 
Such a mapping method may increase classification error.
However, the monotonicity of the mapping curve guarantees that the mapping will not alter the classification results a lot.

In the post-training phase, nonlinear mapping can be applied to specific layers of a model to improve model nonlinearity, which in turn can enhance its resistance to adversarial attacks. 
We believe that nonlinearity is more crucial in the last few layers in a model than earlier layers for the following two reasons: 
First, mapping earlier layers will introduce more accuracy loss than mapping latter layers. 
For instance, our initial experiment shows that if the first basic residual block in ResNet-32 for the CIFAR-10 dataset is mapped nonlinearly, the accuracy of clean image will drop from 95\% to around 15\%. 
Comparably, mapping only the last residual block only leads to an accuracy loss of less than 0.1\%. 
Second, earlier layers in a model are mainly to extract features from the inputs, while the last few layers act as classifiers to vote on the extracted features and to make decisions. 
Adversarial perturbations have a larger impact on models’ decision-making than feature extraction, as shown in~\cite{inkawhich2019feature}. 
Defending in the last few layers of a model is more reasonable to mitigate the error introduced by adversarial perturbations.

\section{Experimental Evaluations}
\label{experiment}

We implement our method with two neural network structures on two datasets: a four-layer CNN on MNIST and a wide ResNet-32 on CIFAR-10. 
Experiments are conducted on TensorFlow (v1.8.0)~\cite{abadi2016tensorflow}. 
Most tests and attacks are performed using the CleverHans library (v2.1.0)~\cite{papernot2016technical}. 
Please note that robustness discussed in this section is represented by the overall accuracy against all considered attacks.




For simplicity, we use abbreviations to represent models and attack types. 
For models with different training techniques, Orig. is a vanilla model trained without additional methods, Adv. was adversarially-trained by~\cite{madry2017towards}, and F.L. was retrained from the original model using feedback learning~\cite{8999302}. 
Models labeled with ``(Q)'' denotes the corresponding weight-quantized models, and ``+mu'' means that $\mu$-law nonlinear mapping is applied after training. 
For adversarial and non-adversarial attacks, we consider Carlini \& Wagner (CW-L2)~\cite{carlini2017towards}, fast gradient sign method (FGSM)~\cite{goodfellow2014explaining}, projected gradient descent (PGD)~\cite{madry2017towards}, basic iterative method (BIM)~\cite{Kurakin2017ICLR}, momentum iterative method (MIM)~\cite{dong2017boosting}, random noise (with a uniform distribution), and Gaussian noise. 
FGSM (w/m/s) represents weak, medium, or strong attack strengths, respectively. 
Note that the attack strengths of attacks vary, as iterative and non-iterative attacks, black-box and white-box attacks all have different degrees of effects on models.

In this section, we only consider the weight quantization and fix the quantization level to 3-bit precision. 
We adopt the same parameters from~\cite{8999302} for the boundary-based error-tolerant retraining.
We select $\mu=255.0$ for nonlinear mapping and apply the $\mu$-law mapping to the last layer of the MNIST model and the last three layers (a residual block and the FC layer) of the CIFAR-10 model.

\subsection{Method Effectiveness}
\label{accuracy}

\begin{table*}[ht]
\centering
\caption{The accuracy of white-box attacks on MNIST models.}
\label{tab:MNIST_white}
\resizebox{.8\textwidth}{!}{%
\begin{tabular}{lrrrrrrr}
\toprule
Models      & Clean            & CW-L2            & FGSM (w)         & FGSM (s)         & PGD              & BIM              & MIM              \\ \cmidrule(lr){1-1} \cmidrule(lr){2-8}
Orig.        & 99.17\%          & 39.40\%          & 73.53\%          & 7.67\%           & 4.38\%           & 5.68\%           & 6.77\%           \\
Orig. (Q)    & 98.97\%          & 36.98\%          & 68.70\%          & 7.40\%           & 2.63\%           & 3.53\%           & 4.27\%           \\ \cmidrule(lr){1-1} \cmidrule(lr){2-8}
Adv.          & 98.40\%          & 94.51\%          & 98.01\%          & 96.24\%          & 97.77\%          & 97.41\%          & 97.32\%          \\
Adv. (Q)      & 42.69\%          & 25.56\%          & 37.28\%          & 32.28\%          & 33.78\%          & 31.44\%          & 30.72\%          \\ \cmidrule(lr){1-1} \cmidrule(lr){2-8}
F.L.        & 99.17\%          & 51.60\%          & 89.69\%          & 39.43\%          & 39.92\%          & 41.42\%          & 43.25\%          \\
F.L. (Q)    & \textbf{98.99\%} & \textbf{49.49\%} & 87.93\%          & 38.36\%          & 35.35\%          & 36.48\%          & 38.33\%          \\ \midrule
Orig.+mu     & 99.06\%          & 34.97\%          & 78.55\%          & 6.32\%           & 7.25\%           & 8.61\%           & 9.04\%           \\
Orig.+mu (Q) & 98.94\%          & 33.09\%          & 73.78\%          & 5.95\%           & 5.21\%           & 6.32\%           & 6.82\%           \\ \cmidrule(lr){1-1} \cmidrule(lr){2-8}
Adv.+mu       & 97.97\%          & 91.77\%          & 97.00\%          & 95.18\%          & 96.79\%          & 95.99\%          & 95.90\%          \\
Adv.+mu (Q)   & 37.12\%          & 28.20\%          & 35.35\%          & 31.15\%          & 34.29\%          & 32.64\%          & 32.15\%          \\ \cmidrule(lr){1-1} \cmidrule(lr){2-8}
F.L.+mu     & 99.11\%          & 48.08\%          & 89.25\%          & 70.86\%          & 57.39\%          & 64.53\%          & 64.92\%          \\
F.L.+mu (Q) & 98.93\%          & 47.65\%          & \textbf{88.31\%} & \textbf{69.45\%} & \textbf{55.24\%} & \textbf{62.64\%} & \textbf{62.92\%} \\ \bottomrule
\end{tabular}%
}
\end{table*}

The comparison for white-box attacks on MNIST is summarized in Table~\ref{tab:MNIST_white}. 
The results show that the Orig. and F.L. models greatly suffer from the adversarial loss, while F.L. performs relatively better than the Orig. model.
The quantization loss affects significantly the Adv. model, which presents a big accuracy gap between the full-precision model and the corresponding quantized version.
Nonlinear mapping does not have much effect on the Orig. and Adv. models. 
Combining the nonlinear mapping with the F.L. model, however, makes a strike on the performance improvements. 

\begin{table*}[ht]
\centering
\caption{The accuracy of black-box attacks and noises on MNIST models.}
\label{tab:MNIST_black}
\resizebox{.8\textwidth}{!}{%
\begin{tabular}{lrrrrrr}
\toprule
Models      & CW-L2            & FGSM (w)         & FGSM (m)         & FGSM (s)         & Normal           & Uniform          \\ \toprule
Orig.        & 97.56\%          & 98.95\%          & 97.80\%          & 93.30\%          & 97.19\%          & 98.85\%          \\
Orig. (Q)    & 97.47\%          & 98.47\%          & 96.26\%          & 90.08\%          & 95.50\%          & 98.38\%          \\ \cmidrule(lr){1-1} \cmidrule(lr){2-7}
Adv.          & 97.28\%          & 98.30\%          & 98.22\%          & 96.17\%          & 77.16\%          & 98.37\%          \\
Adv. (Q)      & 39.42\%          & 45.09\%          & 43.14\%          & 28.02\%          & 17.62\%          & 42.99\%          \\ \cmidrule(lr){1-1} \cmidrule(lr){2-7}
F.L.        & 97.04\%          & 98.90\%          & 97.36\%          & 94.99\%          & 97.01\%          & 98.67\%          \\
F.L. (Q)    & 96.38\%          & \textbf{98.54\%} & \textbf{96.84\%} & \textbf{94.38\%} & \textbf{96.58\%} & \textbf{98.44\%} \\ \midrule
Orig.+mu     & 97.31\%          & 98.72\%          & 97.16\%          & 90.61\%          & 96.16\%          & 98.69\%          \\
Orig.+mu (Q) & 96.83\%          & 98.31\%          & 96.15\%          & 88.69\%          & 95.16\%          & 98.27\%          \\ \cmidrule(lr){1-1} \cmidrule(lr){2-7}
Adv.+mu       & 97.44\%          & 97.83\%          & 97.62\%          & 94.09\%          & 74.06\%          & 97.81\%          \\
Adv.+mu (Q)   & 38.02\%          & 40.06\%          & 39.60\%          & 24.48\%          & 15.69\%          & 37.32\%          \\ \cmidrule(lr){1-1} \cmidrule(lr){2-7}
F.L.+mu     & 97.47\%          & 98.70\%          & 96.72\%          & 93.76\%          & 96.64\%          & 98.58\%          \\
F.L.+mu (Q) & \textbf{97.68\%} & 98.46\%          & 96.44\%          & 93.54\%          & 96.36\%          & 98.21\%           \\ \bottomrule
\end{tabular}%
}
\end{table*}

For black-box attacks on MNIST as shown in Table~\ref{tab:MNIST_black}, all full-precision models have high accuracy. 
Compared with the results in Table~\ref{tab:MNIST_white}, white-box attacks have more severe results than black-box attacks with the same attack strengths. 
This difference is mainly due to the transferability issue as mentioned in \textit{Background}. 
All quantized models except adversarially-trained ones maintain high accuracy; among them, the models with the best performance are the F.L. models, with or without the nonlinear mapping.

\begin{table*}[ht]
\centering
\caption{The accuracy of white-box attacks on CIFAR-10 models.}
\label{tab:CIFAR_white}
\resizebox{.8\textwidth}{!}{%
\begin{tabular}{lrrrrrrr}
\toprule
Models      & Clean                       & CW-L2                       & FGSM (w)                    & FGSM (s)                    & PGD                         & BIM                         & MIM                         \\ \cmidrule(lr){1-1} \cmidrule(lr){2-8}
Orig.        & 95.00\%                     & 9.30\%                      & 20.90\%                     & 10.60\%                     & 2.20\%  & 2.60\%                      & 2.50\%                      \\
Orig. (Q)    & 47.92\% & 13.60\%                     & 16.80\%                     & 11.90\% & 11.10\%                     & 17.80\%                     & 17.70\%                     \\ \cmidrule(lr){1-1} \cmidrule(lr){2-8}
Adv.          & 87.27\%                     & 54.20\%                     & 74.70\%                     & 36.80\%                     & 66.80\%                     & 57.60\%                     & 59.70\%                     \\
Adv. (Q)      & 19.84\%                     & 15.80\%                     & 17.50\%                     & 10.90\%                     & 17.90\%                     & 18.20\%                     & 17.70\%                     \\ \cmidrule(lr){1-1} \cmidrule(lr){2-8}
F.L.        & 93.77\%                     & 20.30\%                     & 39.70\%                     & 27.50\%                     & 4.00\%                      & 4.00\%                      & 4.00\%                      \\
F.L. (Q)    & 90.14\%                     & 21.30\%                     & 42.60\%                     & 28.70\%                     & 5.90\%                      & 5.90\%                      & 5.80\%                      \\ \midrule
Orig.+mu     & 94.05\%                     & 5.30\%  & 95.30\% & 94.90\% & 64.40\% & 95.30\% & 95.30\% \\
Orig.+mu (Q) & 51.55\% & 11.60\% & 45.10\% & 46.80\% & 30.80\% & 49.50\% & 49.40\% \\ \cmidrule(lr){1-1} \cmidrule(lr){2-8}
Adv.+mu       & 85.70\% & 51.90\%                     & 83.30\%                     & 83.20\%                     & 81.60\%                     & 83.30\%                     & 83.30\%                     \\
Adv.+mu (Q)   & 16.80\% & 17.00\% & 16.70\% & 16.70\% & 17.00\% & 17.30\% & 17.50\% \\ \cmidrule(lr){1-1} \cmidrule(lr){2-8}
F.L.+mu     & 93.80\% & 20.70\% & 92.80\% & 92.30\% & 89.50\% & 92.80\% & 92.80\% \\
F.L.+mu (Q) & \textbf{92.20\%} & \textbf{23.10\%} & \textbf{90.80\%} & \textbf{90.70\%} & \textbf{86.90\%} & \textbf{90.80\%} & \textbf{90.80\%} \\ \bottomrule
\end{tabular}%
}
\end{table*}

As the CIFAR-10 dataset is more complicated than MNIST, we implement a model with a much deeper structure. 
Although the results have a broader range than the MNIST results, the observations and conclusions remain consistent. 
As shown in Table~\ref{tab:CIFAR_white}, the Orig. model suffers a lot from adversarial loss, and the Adv. model is still the one that has a considerable accuracy drop after quantization. 
The quantized F.L. model with nonlinear mapping outperforms all other quantized models in our white-box attack scenario to the CIFAR-10 dataset.

\begin{table*}[t]
\centering
\caption{The accuracy of black-box attacks and noises on CIFAR-10 models.}
\label{tab:CIFAR_black}
\resizebox{.8\textwidth}{!}{%
\begin{tabular}{lrrrrrr}
\toprule
Models      & CW-L2            & FGSM (w)         & FGSM (m)         & FGSM (s)         & Normal           & Uniform          \\ \toprule
Orig.        & 58.90\%          & 55.07\%          & 46.87\%          & 41.12\%          & 21.40\%          & 43.80\%          \\
Orig. (Q)    & 23.00\%          & 22.60\%          & 20.64\%          & 19.17\%          & 19.30\%          & 21.80\%          \\ \cmidrule(lr){1-1} \cmidrule(lr){2-7}
Adv.          & 76.44\%          & 75.82\%          & 74.61\%          & 73.48\%          & 70.30\%          & 84.90\%          \\
Adv. (Q)      & 19.38\%          & 19.32\%          & 18.92\%          & 18.55\%          & 15.60\%          & 17.80\%          \\ \cmidrule(lr){1-1} \cmidrule(lr){2-7}
F.L.        & 64.70\%          & 61.82\%          & 57.12\%          & 53.68\%          & 79.10\%          & 85.50\%          \\
F.L. (Q)    & \textbf{62.99\%} & \textbf{60.30\%} & \textbf{56.07\%} & \textbf{52.44\%} & \textbf{72.40\%} & \textbf{81.90\%} \\ \midrule
Orig.+mu     & 55.95\%          & 52.58\%          & 44.74\%          & 38.62\%          & 20.90\%          & 41.00\%          \\
Orig.+mu (Q) & 25.64\%          & 24.53\%          & 21.41\%          & 20.01\%          & 15.80\%          & 19.30\%          \\ \cmidrule(lr){1-1} \cmidrule(lr){2-7}
Adv.+mu       & 73.24\%          & 72.79\%          & 71.52\%          & 69.90\%          & 68.20\%          & 82.30\%          \\
Adv.+mu (Q)   & 15.74\%          & 15.67\%          & 15.23\%          & 14.69\%          & 11.10\%          & 12.10\%          \\ \cmidrule(lr){1-1} \cmidrule(lr){2-7}
F.L.+mu     & 63.69\%          & 60.37\%          & 55.58\%          & 52.04\%          & 73.60\%          & 84.00\%          \\
F.L.+mu (Q) & 62.54\%          & 59.65\%          & 55.03\%          & 51.69\%          & 72.20\%          & \textbf{81.90\%} \\ \bottomrule
\end{tabular}%
}
\end{table*}

For black-box attacks on CIFAR-10 in Table~\ref{tab:CIFAR_black}, the nonlinear mapping does not help and it has a minor negative effect on the models' performance. 
We believe the accuracy drop after applying nonlinear mapping is negligible, compared to the accuracy recovery in white-box gradient-based attacks.

Overall, the best quantized model for defending both white-box and black-box adversarial attacks is F.L.+mu (Q), which has an average improvement of 20.54\% on MNIST and 52.69\% on CIFAR-10 compared to the Orig. model. Compared to the second best model, F.L. (Q), our method's average accuracy improves 7.55\% on MNIST and 27.84\% on CIFAR-10.




\subsection{Verification with the Lipschitz Constant}

\begin{figure*}[ht]
	\centering
	\includegraphics[width=0.8\linewidth]{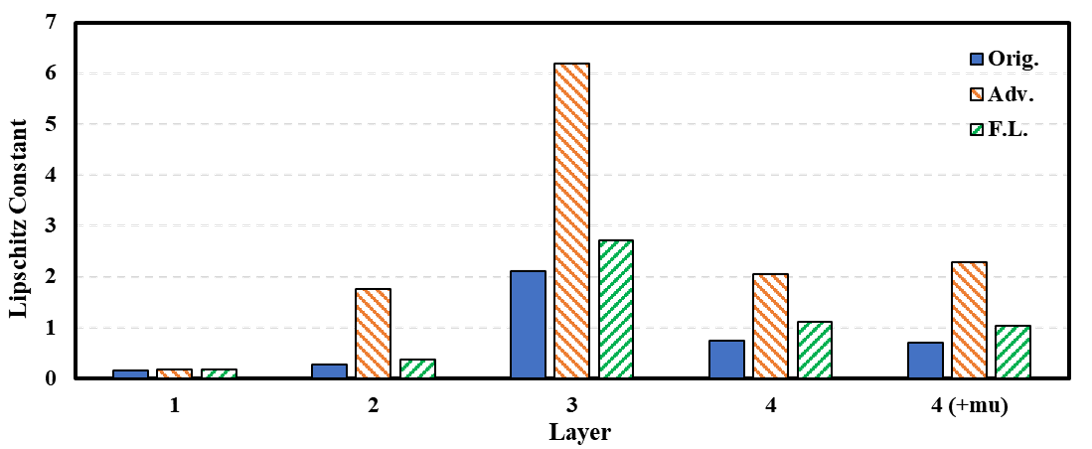}
	\caption{The Lipschitz constant of $\Delta W$ in different layers (MNIST).}
	\label{fig:Lipschitz_MNIST}
\end{figure*}

\begin{figure*}[ht]
	\centering
	\includegraphics[width=0.8\linewidth]{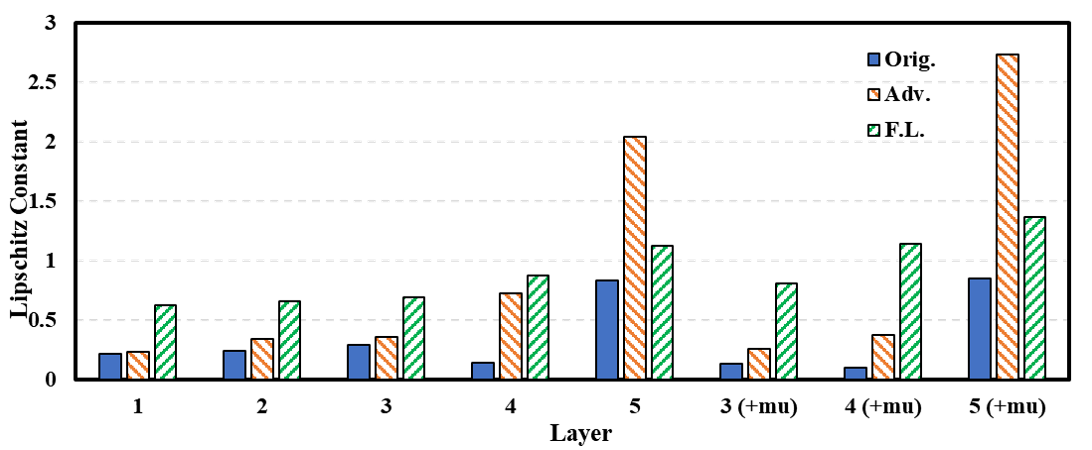}
	\caption{The Lipschitz constant of $\Delta W$ in different layers (CIFAR-10).}
	\label{fig:Lipschitz_CIFAR}
\end{figure*}

To further validate the effectiveness of our method, we examine the Lipschitz constant of $\Delta W$ of some layers. 
Here we only consider $p=2$ for simplicity. 
As given in Eq.~(\ref{eq:2}), when $p=2$, the Lipschitz constant of $\Delta W$ is the maximum singular value of $\Delta W$. The layer-wise results for MNIST and CIFAR-10 are shown in Figure~\ref{fig:Lipschitz_MNIST} and Figure~\ref{fig:Lipschitz_CIFAR}, respectively.

In Figure~\ref{fig:Lipschitz_MNIST}, the $x$-axis represents the number of layers. In Figure~\ref{fig:Lipschitz_CIFAR}, layers 1 to 4 denote the last two residual blocks, and layer 5 is the last FC layer. 
The ``(+mu)" in both figures represents the specific layers are mapped nonlinearly. 
We only map the last layer for the MNIST dataset, and the last three layers (one residual block with two layers and the only FC layer) for the CIFAR-10 dataset. 
Our findings from the results are summarized as follows:
\begin{itemize}[leftmargin=*]
	\item The Adv. model has the largest Lipschitz results among three models on MNIST, the Orig. model has the lowest, and the F.L. model is slightly worse than the Orig. model.
	\item For CIFAR-10, the results are varied by layers. For all layers except the last layer, the Orig. and Adv. models both have lower Lipschitz constants than the F.L. model's Lipschitz constants, which are still less than 1 and only have minor quantization error amplification effect. For the last layer, however, Adv. has a Lipschitz constant over 2, greater than the other two models.
	\item Using nonlinear mapping will not significantly change the Lipschitz constant of $\Delta W$ in the Orig. and F.L. models. For the last layer in the CIFAR-10 model, Adv. result increases by 34.29\%, far larger than Orig. (2.28\%) and Adv. (22.20\%) results.
\end{itemize}

The Lipschitz results show that adversarial training has weak tolerance to quantization, which is consistent with Figure~\ref{fig:motivation}.
The results also reveal that adversarial training is incapable of cooperating with other techniques, such as the nonlinear mapping method. 
A Lipschitz constant of $\Delta W$ greater than 1 means the quantization error may be amplified in this layer. 
Based on the results shown in \textit{Method Effectiveness} and the Lipschitz results, we believe the last few layers, especially the last layer, matter more than the other layers in terms of the Lipschitz constant of $\Delta W$. Such results indicate that the last few layers have more notable quantization error amplification effect than the previous layers. 

\subsection{Other Experiment Attempts}

We also implement other training techniques such as the quantization-aware training and the defensive quantization method proposed in~\cite{lin2019defensive}, but the results are not as good as expected or reported. 
Combining the adversarial training with the quantization-aware training has even worse performance than implementing only adversarial training (results are omitted due to page limit). 
The downgrade of such a combination is because the goals of adversarial training and quantization-aware training are not well aligned with each other. 
The results also indicate the importance and necessity of an integrated method to mitigate both adversarial and quantization losses. 
Another attempt is to replicate~\cite{lin2019defensive}. 
The defensive quantization method is simple to understand and implement, but the computation cost of the regularization term becomes too much as the neural network gets deeper and wider. Therefore these methods are not considered as our baselines.
\section{Conclusions}
\label{conclusions}

This work focuses on the robustness of quantized neural networks. 
We observe that adversarially-trained neural networks are vulnerable to quantization loss.
For the reason, adversarial training cannot be directly applied to quantized models. 
In this work, we theoretically analyze both adversarial and quantization losses and come up with criteria to measure the two losses. 
We also propose a solution to minimize both losses at the same time and conduct various experiments to show its effectiveness. 
The results show that our method is capable of defending both black-box and white-box gradient-based adversarial attacks and minimizing the quantization loss, showing an average accuracy improvement against adversarial attacks of 7.55\% on MNIST and 27.84\% on CIFAR-10 compared to the next best approach studied.

\bibliography{AAAI21.bib}

\end{document}